\documentclass{article}

\usepackage{microtype}
\usepackage{graphicx}
\usepackage{subcaption}
\usepackage{booktabs}
\usepackage{hyperref}

\newif\ifarxiv
\arxivtrue

\ifarxiv
  \usepackage[preprint]{icml2026}
  \usepackage{etoolbox}
  \makeatletter
  \renewcommand{\Notice@String}{}
  \patchcmd{\printAffiliationsAndNotice}{\par\ \\ \Notice@String}{}{}%
    {\PackageWarning{main}{Affiliation footnote patch failed}}
  \makeatother
  \hypersetup{pdfsubject={}}
\else
  \usepackage{icml2026}
\fi

\usepackage[utf8]{inputenc} 
\usepackage[T1]{fontenc}    
\usepackage{amsmath}
\usepackage{bm}
\usepackage{amssymb}
\usepackage{amsfonts}       
\usepackage{mathtools}
\usepackage{amsthm}
\usepackage{nicefrac}       
\usepackage{microtype}      
\usepackage{xcolor}
\usepackage{colortbl}
\usepackage{booktabs}       
\usepackage{multicol}
\usepackage{multirow}
\usepackage{wrapfig}
\usepackage{graphicx}
\usepackage{float}          
\usepackage{pifont}
\usepackage{enumitem}       
\usepackage{tikz}
\usetikzlibrary{decorations.pathreplacing}
\usepackage[capitalize,noabbrev,nameinlink]{cleveref}
\creflabelformat{equation}{#2#1#3}

\theoremstyle{plain}

\theoremstyle{definition}

\theoremstyle{remark}

\usepackage{xurl}
\definecolor{linkindigo}{HTML}{1A13AB}
\hypersetup{
    colorlinks=true,
    linkcolor=linkindigo,
    citecolor=linkindigo,
    filecolor=magenta,
    urlcolor=linkindigo,
}

\newcommand{\panelref}[2]{\hyperref[#1]{\cref*{#1}#2}}

\renewcommand{\paragraph}[1]{\par\vspace{1.5pt}\noindent\textbf{#1}\hspace{0.5em}\ignorespaces}

\makeatletter
\g@addto@macro\normalsize{%
  \abovedisplayskip 6\p@
  \belowdisplayskip 6\p@
  \abovedisplayshortskip \z@
  \belowdisplayshortskip 4\p@}
\makeatother

\definecolor{planblue}{HTML}{0070C0}
\definecolor{refineamber}{HTML}{D8A300}
\definecolor{rendergreen}{HTML}{3B7D23}

\newcommand{\shline}{\noalign{\vspace{1pt}}\hline\noalign{\vspace{1pt}}}
\newcommand{\shlineshaded}{\noalign{\vspace{1pt}}\hline\noalign{{\color{baselinecolor}\hrule height 1pt}}\rowcolor{baselinecolor}}
\newcommand{\shlinefromshaded}{\noalign{{\color{baselinecolor}\hrule height 1pt}}\hline\noalign{\vspace{1pt}}}
\makeatletter
\newcommand{\gaprow}[1]{\noalign{\global\setbox\@arstrutbox\hbox{\vrule\@height1pt\@depth\z@\@width\z@}}#1\\\noalign{\global\setbox\@arstrutbox\hbox{\vrule\@height\arraystretch\ht\strutbox\@depth\arraystretch\dp\strutbox\@width\z@}}}
\newcommand{\tablestyle}[2]{\setlength{\tabcolsep}{#1}\centering\footnotesize
  \dimen@=#2\dimexpr\ht\strutbox+\dp\strutbox\relax
  \setbox\strutbox\hbox{\vrule\@height\dimexpr(\dimen@+\fontcharht\font`0)/2\relax\@depth\dimexpr(\dimen@-\fontcharht\font`0)/2\relax\@width\z@}%
  \renewcommand{\arraystretch}{1}%
  \setlength{\aboverulesep}{1pt}\setlength{\belowrulesep}{1pt}%
  \setlength{\heavyrulewidth}{\arrayrulewidth}\setlength{\lightrulewidth}{\arrayrulewidth}}
\newlength{\rulepad}
\newcommand{\padabove}{\rule{0pt}{\dimexpr\ht\strutbox+\rulepad\relax}}
\newcommand{\padbelow}{\rule[-\dimexpr\dp\strutbox+\rulepad\relax]{0pt}{0pt}}
\newcommand{\headspan}[1]{\multirow{2}{*}[-\dimexpr\rulepad+(\aboverulesep+\cmidrulewidth+\belowrulesep)/2\relax]{#1}}
\makeatother

\colorlet{baselinecolor}{gray!10}

\definecolor{worsered}{HTML}{C0392B}
\newcommand{\worse}[1]{\textcolor{worsered}{#1}}
\definecolor{bettergreen}{HTML}{229954}
\newcommand{\better}[1]{\textcolor{bettergreen}{#1}}

\newcommand{\method}{LoopRes}

\definecolor{projectlinkblue}{HTML}{3B6EA8}
\newcommand{\projectlink}[3]{{\hypersetup{urlcolor=projectlinkblue}\href{#1}{\raisebox{-0.2em}{\includegraphics[height=1.15em]{figures/icons/#2}}\hspace{0.4em}{\fontfamily{txtt}\selectfont #3}}}}

\usepackage{listings}
\makeatletter
\let\old@lstKV@SwitchCases\lstKV@SwitchCases
\def\lstKV@SwitchCases#1#2#3{}
\makeatother
\usepackage{lstlinebgrd}
\makeatletter
\let\lstKV@SwitchCases\old@lstKV@SwitchCases
\lst@Key{numbers}{none}{%
  \def\lst@PlaceNumber{\lst@linebgrd}%
  \lstKV@SwitchCases{#1}%
  {none:\\%
   left:\def\lst@PlaceNumber{\lst@linebgrd\llap{\normalfont\lst@numberstyle{\thelstnumber}\kern\lst@numbersep}}\\%
   right:\def\lst@PlaceNumber{\lst@linebgrd\rlap{\normalfont\kern\linewidth\kern\lst@numbersep
     \lst@numberstyle{\thelstnumber}}}%
  }{\PackageError{Listings}{Numbers #1 unknown}\@ehc}}
\makeatother
\definecolor{codenum}{HTML}{A0A7B1}
\definecolor{codetext}{HTML}{24292F}
\definecolor{codekw}{HTML}{CF222E}
\definecolor{codefn}{HTML}{8250DF}
\definecolor{codestr}{HTML}{0A3069}
\definecolor{codecmt}{HTML}{6E7781}
\definecolor{codebuiltin}{HTML}{0550AE}
\definecolor{codehl}{HTML}{DCEBF7}
\makeatletter
\newcommand{\codelinebg}[3]{\raisebox{0pt}[0pt][0pt]{\color@block{#1}{#2}{#3}}}
\makeatother
\lstdefinestyle{pycode}{
  language=Python,
  basicstyle=\fontfamily{txtt}\selectfont\fontsize{6.7pt}{9pt}\selectfont\color{codetext},
  columns=fullflexible, keepspaces=true, literate={-}{-}1, upquote=true, showstringspaces=false,
  numbers=left, numberstyle=\fontfamily{txtt}\selectfont\fontsize{5.5pt}{9pt}\selectfont\color{codenum},
  numbersep=8pt, xleftmargin=2.2em, framexleftmargin=2.2em, framexrightmargin=0pt,
  linebackgroundsep=2.2em, linebackgroundwidth=\dimexpr\linewidth+2.2em\relax,
  linebackgroundheight=6.6pt, linebackgrounddepth=2.6pt, linebackgroundcmd=\codelinebg,
  frame=tb, framerule=\arrayrulewidth, rulecolor=\color{black},
  framesep=4pt, aboveskip=0pt, belowskip=0pt,
  keywordstyle=\bfseries\color{codekw}, morekeywords={self,None,True,False,del},
  keywordstyle=[2]\color{codebuiltin}, morekeywords=[2]{torch,len,range,enumerate,list,Tensor,RMSNorm},
  deletecomment=[s]{"""}{"""}, morestring=[s]{"""}{"""},
  stringstyle=\color{codestr}, commentstyle=\itshape\color{codecmt},
  emph={loop_res,forward}, emphstyle=\bfseries\color{codefn},
}

\def\eqref#1{equation~\ref{#1}}

\def\1{\bm{1}}

\def\vb{{\bm{b}}}

\def\ve{{\bm{e}}}

\def\vh{{\bm{h}}}

\def\vk{{\bm{k}}}

\def\vp{{\bm{p}}}
\def\vq{{\bm{q}}}

\def\vu{{\bm{u}}}
\def\vv{{\bm{v}}}
\def\vw{{\bm{w}}}
\def\vx{{\bm{x}}}

\def\vz{{\bm{z}}}

\def\mI{{\bm{I}}}

\DeclareMathAlphabet{\mathsfit}{\encodingdefault}{\sfdefault}{m}{sl}
\SetMathAlphabet{\mathsfit}{bold}{\encodingdefault}{\sfdefault}{bx}{n}

\icmltitlerunning{Unlocking the Power of Loops in Pretrained LLMs with Loop Residuals}

\begin{document}

\twocolumn[
  \icmltitle{Unlocking the Power of Loops in Pretrained LLMs with Loop Residuals}

  \icmlsetsymbol{equal}{*}

  \begin{icmlauthorlist}
    \icmlauthor{Xuanlei Zhao}{nus}
    \icmlauthor{Hailun Xu}{}
    \icmlauthor{Kai Wang}{nus}
    \icmlauthor{Yang Luo}{nus}
    \icmlauthor{Yingxuan Ren}{nus}
    \icmlauthor{Zirui Zhu}{nus}
    \icmlauthor{Yuxuan Lou}{nus}
    \icmlauthor{Kun Xu}{}
    \icmlauthor{Yan Wang}{ind}
    \icmlauthor{Kanchan Sarkar}{}
    \icmlauthor{Yang You}{nus}
  \end{icmlauthorlist}

  \ifarxiv
    \vskip 0.04in
    {\centering\normalsize
      \projectlink{https://github.com/NUS-HPC-AI-Lab/LoopRes}{github}{github.com/NUS-HPC-AI-Lab/LoopRes}\par}
  \fi

  \icmlaffiliation{nus}{National University of Singapore}
  \icmlaffiliation{ind}{Independent Researcher}
  \icmlcorrespondingauthor{Xuanlei Zhao}{xuanlei.zhao@u.nus.edu}
  \icmlcorrespondingauthor{Yang You}{yangyou@nus.edu.sg}

  \icmlkeywords{Looped Transformers, Machine Learning, ICML}

  \vskip 0.3in

\begin{center}
  \vspace{-5pt}
  \captionsetup{type=figure}
  \subcaptionbox{1.7B, 2$\times$ and 4$\times$ compute.\label{fig:teaser_avg}}{%
    \includegraphics{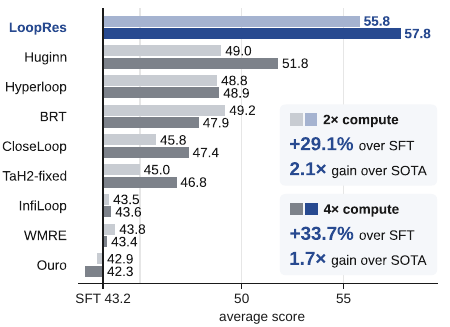}}\hfill
  \subcaptionbox{1.7B scaling, 2$\times$ compute.\label{fig:teaser_small}}{%
    \includegraphics{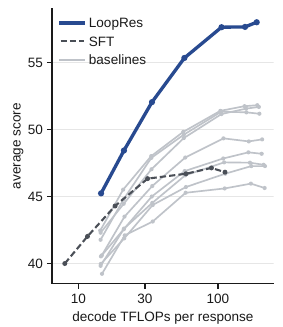}}\hfill
  \subcaptionbox{4B/8B scaling, 2$\times$ compute.\label{fig:teaser_large}}{%
    \includegraphics{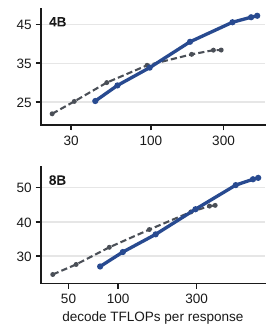}}
  \vspace{-3pt}
  \caption{\textbf{Average scores across compute budgets and backbones.}
    Average scores on seven math and code benchmarks, where SFT is the same model finetuned without looping.
    (a) Qwen3-1.7B-Base at 2$\times$ (light) and 4$\times$ (dark) compute, with SOTA the best baseline at the same compute.
    (b) Scores against decode compute as the thinking budget grows from 1K to 30K tokens.
    (c) The same sweep on Qwen3-4B-Base and Qwen3-8B-Base over five math and code benchmarks.}
  \label{fig:teaser}
\end{center}

  \vskip 0.2in
]

\printAffiliationsAndNotice{\raggedright}

\begin{abstract}
Looped Transformers iterate a shared block of layers to increase computation
without adding parameters, offering stronger performance and test-time
scaling.
However, this power has so far required training from scratch and yielded
only modest gains for LLMs.
Finetuning existing models into loops is far cheaper, yet also brings only
marginal gains over SFT.
We propose \method{}, an effective, scalable, and simple approach that
turns loops into significant and consistent gains for pretrained LLMs.
Building on Attention Residuals, \method{} lets each looped layer attend
directly to the outputs of earlier loops.
Since these outputs are iterates of the same layers and become harder to
tell apart, we give each a trainable loop-aware key that encodes its
structural identity.
Attending to all earlier loops also lets layers skip intermediate ones, so
we restrict each layer to a residual window of the latest $W$ loops,
forcing every loop to build incrementally on recent ones.
Across 6 pretrained LLMs from 3 model families and 9 benchmarks, \method{}
delivers significant and consistent gains: at 2$\times$ compute, it
improves Qwen3-1.7B-Base by 29.1\% over SFT, more than doubling the gain
of the state-of-the-art baseline.
The gains hold on larger models up to 30B and across all three families.
More importantly, from 1.7B to 8B, \method{} scales more steeply than SFT
as compute grows, reaching a higher ceiling while needing less compute on
long outputs.
These results establish \method{} as an effective looping approach,
delivering both better performance and better test-time scaling.
\end{abstract}

\section{Introduction}
\label{sec:intro}

\begin{figure*}[t]
  \centering
  \includegraphics[width=\textwidth]{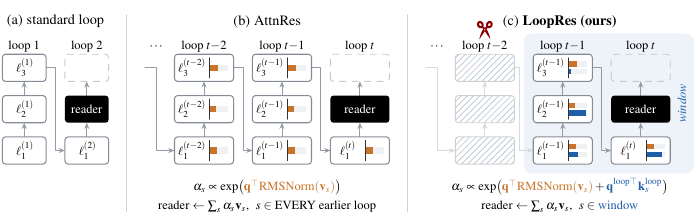}
  \vspace{-20pt}
  \caption{\textbf{How a looped layer takes the outputs of earlier layers as input.}
    Each panel diagrams three layers running for up to three loops, highlighting one layer's input, where $\smash{\ell^{(t)}_i}$ denotes layer $i$ of loop $t$ and dashed layers have not yet run. Loops before $t{-}2$ are omitted in (b) and (c).
    (a)~A standard loop feeds only the last layer of each loop into the first of the next.
    (b)~AttnRes attends to every earlier loop output with content-based weights.
    (c)~\method{} attends only to a window of the latest $W$ loops (shaded, $W{=}1$ here) and adds a loop-aware key alongside the content key. Each bar marks the magnitude of a score term.}
  \label{fig:grid}
  \vspace{-5pt}
\end{figure*}

Looped Transformers reuse a shared block of layers multiple times in a single forward pass, achieving deeper computation without adding parameters~\cite{dehghani2019universal,giannou2023looped}.
Depth scales with the number of loops rather than the number of layers, making loops a simple lever for both stronger performance and test-time compute scaling.
Yet turning this promise into significant and consistent gains for LLMs remains an open challenge.

One line of work pretrains looped LLMs from scratch~\cite{geiping2025huginn,zhu2025ouro}.
However, the gains remain modest or even negative at comparable cost~\cite{zeitoun2026hyperloop,wang2026smelt}.
Given the expense of pretraining, this direction offers limited returns.

A more promising approach finetunes pretrained LLMs into loops~\cite{marchenko2026closing,you2026tah2}, reusing their learned capabilities at a fraction of the pretraining cost.
As illustrated in \panelref{fig:grid}{a}, although these designs vary in their looping pattern~\cite{zhu2025ouro,geiping2025huginn,you2026tah2,zeitoun2026hyperloop,wang2026smelt,zhuang2026recurrence,marques2026brt}, they all compress the outputs of earlier loops into a narrow pathway.
This limits what later loops can build on, yielding only marginal gains and leaving the potential of loops largely unrealized.

A natural remedy is to let each looped layer attend directly to the outputs of earlier loops.
Attention Residuals (AttnRes)~\cite{kimi2026attnres} offer exactly this: each layer builds its input by attending to all earlier layer outputs.
However, AttnRes is designed for standard Transformers, where the sources differ in content and are fixed in number.
Our key insight is that loops break both assumptions: the sources are iterates of the same layers, so they become similar in content and grow in number.

This causes three problems.
(i) \textbf{Similar iterates.}
The outputs are iterates of the same layers and grow harder to tell apart as loops deepen.
(ii) \textbf{Early convergence.}
The AttnRes weights converge early in finetuning, and because the keys come from similar outputs, the model cannot learn to tell sources apart before the pattern is set.
(iii) \textbf{Growing sources.}
Each loop adds new sources to the full attention window, so layers can shortcut to earlier loops, skipping intermediate ones.

Based on these observations, we propose Loop Residuals (\method{}), an effective, scalable, and simple approach that turns loops into significant and consistent gains for pretrained LLMs.
\method{} builds on AttnRes and introduces two improvements.
(i) It assigns each source a trainable, loop-aware key alongside its content key, encoding the structural relationship of loops and layers.
Because these keys are directly optimized, the model can learn to differentiate sources before the weights converge.
(ii) It restricts each layer to a residual window of the latest $W$ loops.
This eliminates shortcuts to earlier loops, forces every loop to build incrementally on recent ones, and caps the overhead.

\begin{figure*}[t]
  \centering
  \includegraphics[width=\textwidth]{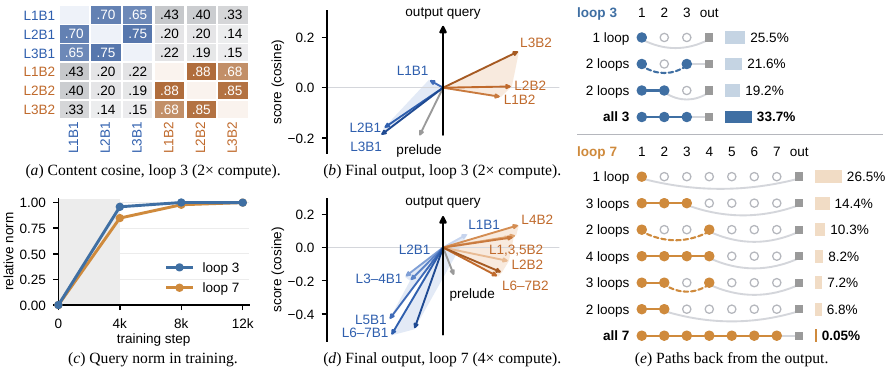}
  \vspace{-20pt}
  \caption{\textbf{Visualization of AttnRes in a looped model.}
    Huginn + AttnRes on Qwen3-1.7B-Base at 2$\times$ compute ($L{=}3$) and 4$\times$ compute ($L{=}7$). L$t$B$j$ denotes the output of block $j$ in loop $t$.
    (a)~Cosine similarity between block outputs across all loops.
    (b,d)~Attention scores from the output query to each source at $L{=}3$ and $L{=}7$. Arrow height marks the score.
    (c)~Output query norm over finetuning steps, normalized by its converged value.
    (e)~Weight share on each path from the output back to the first loop, obtained by multiplying the attention weights along the path.}
  \label{fig:why}
\end{figure*}

Across 6 pretrained LLMs from 3 model families and 9 math and code benchmarks, \method{} delivers significant and consistent gains.
On Qwen3-1.7B-Base at 2$\times$ compute, it improves the average score by 29.1\% over SFT, more than doubling the gain of the SOTA baseline, and its gain continues to grow as loops deepen.
The gains are consistent: \method{} improves by 10.0\% to 37.8\% on the larger Qwen3-\{4B,8B,30B-A3B\}-Base, as well as on Qwen3.5 and Llama-3.2 from different families.
Notably, \method{} on the 4B backbone even surpasses SFT on the 8B one, showing that loops can bridge the gap to larger models.

More importantly, \method{} achieves better scaling at test time: from 1.7B to 8B, its score rises with a steeper slope than SFT as compute grows.
It thus reaches a higher ceiling, significantly exceeding the best score of SFT.
At the same time, it is more efficient on long outputs, reaching the same score at a lower compute cost.
Together, these results establish loop depth as an axis of test-time scaling, benefiting both inference and post-training rollouts.
We hope \method{} serves as a simple and effective foundation for scaling LLMs along this axis.

\section{Realizing the Promise of Looping}
\label{sec:method}

\subsection{Preliminaries}
\label{sec:prelim}

\begin{figure}[h!]
  \centering
  \includegraphics{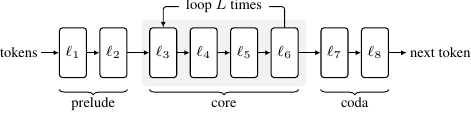}
  \vspace{-5pt}
  \caption{\textbf{The looped architecture of Huginn.}
    The layers of an LLM are split into a prelude, a core, and a coda.
    The prelude and the coda run once, and the core runs $L$ times.}
  \label{fig:huginn}
  \vspace{-5pt}
\end{figure}

\paragraph{Looped language models.}
We adopt the looped architecture of Huginn~\cite{geiping2025huginn} as shown in \cref{fig:huginn} and build it from a pretrained LLM by splitting its layers into three consecutive parts: a prelude, a core, and a coda.
The prelude runs once on the token embeddings and produces $\vp$.
The core then runs $L$ times, and we call each run a loop, indexed by $t \in \{1, \dots, L\}$.
The coda runs once on the output of loop $L$ to predict the next token, and the model is finetuned with the cross-entropy loss on this prediction.

\paragraph{Attention Residuals.}
A standard Transformer feeds each sub-layer the residual stream.
Attention Residuals (AttnRes)~\cite{kimi2026attnres} replace the stream with a learned input.
In its block form, the layers are grouped into blocks, and each sub-layer takes as input a set of sources $\mathcal{S}$: the embedding, the summed output of every completed block, and the partial sum of the current block.
AttnRes weighs these sources by a softmax attention, and the input of the sub-layer is formulated as
\begin{equation}
  \vx = \sum_{s \in \mathcal{S}} \alpha_s \vv_s,
  \qquad
  \alpha_s = \frac{\exp(\vq^\top \vk_s)}{\sum_{s' \in \mathcal{S}} \exp(\vq^\top \vk_{s'})},
  \label{eq:attnres}
\end{equation}
where $\vq$ is a learnable query of the sub-layer initialized to zero, $\vv_s$ is a source, and $\vk_s = \mathrm{RMSNorm}(\vv_s)$ is its key.
We call $\vk_s$ a content key, since it is derived from the source and makes the weights content-dependent.

\subsection{Applying Attention Residuals to Loops}
\label{sec:attnres_loop}

We apply AttnRes only to the core of the looped model.
The prelude and the coda keep the standard residual stream.
Within the core, we group its layers into $B$ blocks.
We denote the output of block $i$ in loop $t$ by $\smash{\vb^{(t)}_{i}}$, and the running partial sum within block $i$ by $\smash{\vu^{(t)}_{i}}$.
A sub-layer in block $i$ of loop $t$ then takes as input
\begin{equation}
  \mathcal{S}^{(t)}_{i} = \{\vp,
    \underbrace{\vb^{(1)}_{1}, \dots, \vb^{(t-1)}_{B}\vphantom{\vb^{(t)}_{i-1}}}_{\text{previous loops}},
    \underbrace{\vb^{(t)}_{1}, \dots, \vb^{(t)}_{i-1}, \vu^{(t)}_{i}}_{\text{loop } t}\},
  \label{eq:sources}
\end{equation}
where $\vp$ is the prelude output.
Each sub-layer forms its input from these sources by \cref{eq:attnres}, using its own query in each loop.
The output of loop $t$ is formed the same way, with its own query.

\subsection{Challenges of Looping with Attention Residuals}
\label{sec:challenges}

\paragraph{Similar iterates.}
As shown in \panelref{fig:why}{a}, the iterates of the same block are far more similar to each other than to other blocks.
Because the content keys are derived from these sources, similar iterates receive similar scores.
With three loops, the output query in \panelref{fig:why}{b} clearly separates the two blocks, but the iterates within each block are hard to tell apart.
With seven loops in \panelref{fig:why}{d}, this worsens, as many iterates collapse onto nearly the same score.

\paragraph{Early convergence.}
As shown in \panelref{fig:why}{c}, the query norm reaches most of its final value within the first third of finetuning and then levels off.
We hypothesize that this commits the model to its early connection pattern, since content keys change only indirectly, through the hidden states.

\paragraph{Growing sources.}
The source set grows with every loop, and AttnRes attends to all of them, so layers can skip intermediate loops.
As shown in \panelref{fig:why}{e}, the weight on the path through every loop falls from 33.7\% with three loops to 0.05\% with seven loops, and the most likely paths never reach the later loops, suggesting that deeper loops add little effective depth.

\subsection{LoopRes}
\label{sec:loopres}

Building on the analysis above, we propose \method{}, an effective, scalable, and simple approach that turns loops into significant and consistent gains for pretrained LLMs.
As illustrated in \panelref{fig:grid}{c}, \method{} keeps the input-dependent mechanism of AttnRes and introduces two improvements: (i) It assigns each source a loop-aware key alongside its content key, which resolves the identity of each iterate. (ii) It restricts each layer to a residual window of the latest $W$ loops, which bounds the number of sources as the loop deepens. \cref{fig:pseudo} provides PyTorch-style pseudo code.

\paragraph{Loop-aware keys.}
We assign each source a loop-aware key $\vk^{\mathrm{loop}}_s$ alongside its content key $\vk_s = \mathrm{RMSNorm}(\vv_s)$: a trainable vector per block per loop, initialized from $\mathcal{N}(\mathbf{0}, \mI)$.
Each sub-layer uses the two keys with two queries, and the input weights in \cref{eq:attnres} become
\begin{equation}
  \alpha_s = \frac{\exp(\vq^{\mathrm{loop}\top}_{\vphantom{s}} \vk^{\mathrm{loop}}_s + \vq^\top \vk_s)}{\sum_{s' \in \mathcal{S}} \exp(\vq^{\mathrm{loop}\top}_{\vphantom{s'}} \vk^{\mathrm{loop}}_{s'} + \vq^\top \vk_{s'})},
  \label{eq:keys}
\end{equation}
where $\vq^{\mathrm{loop}}$ is the loop-aware query and $\vq$ the content query, both initialized to zero, so that each sub-layer starts from uniform weights over its sources.

While the content key determines how relevant each source is to the current token, the loop-aware key brings two benefits: (i) It encodes the structure of the loop, distinguishing the iterates of the same layers by their coordinates. (ii) It is optimized directly, so the connection pattern no longer adapts only through the hidden states.

\paragraph{Residual window.}
\method{} restricts the residual sources to a window of the latest $W$ loops.
This brings three benefits: (i) It prevents layers from skipping to distant loops, forcing each loop to build on the latest loops. (ii) It keeps the softmax from shrinking the weight of the current loop as sources accumulate. (iii) It bounds each input to $(W+1)B + 1$ sources, thus limiting compute, memory, and communication overhead to $W$ loops.

\paragraph{Efficiency.}
A naive implementation of \cref{eq:keys} loads all queries and keys, then writes the weighted sum for every sub-layer.
Following the design of FlashAttention~\cite{dao2024flashattention2}, we fuse these steps into a single Triton~\cite{tillet2019triton} kernel that loads each source once from HBM and writes only the result back, significantly reducing IO cost and peak memory.

\begin{table*}[t]
\caption{\textbf{Comparison with baselines.} All methods run on Qwen3-1.7B-Base at 2$\times$ and 4$\times$ compute, measured relative to SFT. HE+ denotes HumanEval+, and both code benchmarks report the plus tests. \emph{avg.} averages the seven math and code scores. Each $\Delta$ column gives a method's change relative to SFT. Within each compute budget, \textbf{bold} marks the best and \underline{underline} the second best.}
\label{tab:main}
\vspace{-7pt}
\tablestyle{5.98pt}{1.1}
\begin{tabular}{l|ccccc|cc|cc|cc}
\padbelow\headspan{method} & \multicolumn{5}{c|}{math} & \multicolumn{2}{c|}{code} & \headspan{avg.} & \headspan{$\Delta$} & \headspan{loss} & \headspan{$\Delta$} \\
\cmidrule(lr){2-6}\cmidrule(lr){7-8}
\padabove\padbelow& AIME25 & AIME26 & AMC23 & Olympiad & MATH500 & HE+ & MBPP+ & & & & \\
\shline
\padabove\padbelow SFT & 14.6 & 11.7 & 48.6 & 40.7 & 73.4 & 57.3 & 56.2 & 43.2 & -- & 0.312 & -- \\
\shline
\multicolumn{12}{c}{\padabove\padbelow\textbf{2$\times$ compute}} \\
\shline
\padabove Huginn & 19.6 & 14.4 & 54.1 & 46.5 & \underline{80.4} & 64.2 & \underline{64.0} & 49.0 & \better{13.5\%$\uparrow$} & 0.303 & \better{2.8\%$\downarrow$} \\
Ouro & 12.5 & 9.2 & 47.8 & 40.4 & 74.6 & 59.1 & 56.6 & 42.9 & \worse{0.8\%$\downarrow$} & 0.315 & \worse{1.1\%$\uparrow$} \\
Hyperloop & 18.3 & \underline{16.7} & \underline{54.4} & \underline{47.1} & 77.6 & \underline{66.9} & 60.4 & 48.8 & \better{12.9\%$\uparrow$} & 0.303 & \better{2.9\%$\downarrow$} \\
TaH2-fixed & 15.8 & 12.5 & 50.0 & 41.3 & 75.8 & 62.3 & 57.0 & 45.0 & \better{4.0\%$\uparrow$} & 0.306 & \better{1.7\%$\downarrow$} \\
WMRE & 16.2 & 10.6 & 51.2 & 40.7 & 74.0 & 60.1 & 54.1 & 43.8 & \better{1.5\%$\uparrow$} & 0.311 & \better{0.3\%$\downarrow$} \\
CloseLoop & 16.5 & 13.3 & 50.0 & 44.1 & 78.8 & 61.4 & 56.7 & 45.8 & \better{6.0\%$\uparrow$} & 0.309 & \better{1.0\%$\downarrow$} \\
InfiLoop & 14.4 & 10.8 & 49.4 & 42.4 & 73.0 & 58.4 & 56.3 & 43.5 & \better{0.7\%$\uparrow$} & 0.312 & \worse{0.0\%$\uparrow$} \\
\padbelow BRT & \underline{21.7} & 15.4 & 53.6 & 45.0 & 79.0 & 66.6 & 63.2 & \underline{49.2} & \better{13.9\%$\uparrow$} & \underline{0.302} & \better{3.0\%$\downarrow$} \\
\shlineshaded
\padabove\padbelow\textbf{\method{}} & \textbf{27.9} & \textbf{19.4} & \textbf{64.7} & \textbf{53.8} & \textbf{84.0} & \textbf{71.8} & \textbf{68.9} & \textbf{55.8} & \better{\textbf{29.1}\%$\uparrow$} & \textbf{0.292} & \better{\textbf{6.3}\%$\downarrow$} \\
\shlinefromshaded
\multicolumn{12}{c}{\padabove\padbelow\textbf{4$\times$ compute}} \\
\shline
\padabove Huginn & \underline{24.4} & \underline{18.3} & \underline{58.8} & \underline{50.4} & \underline{80.0} & \underline{66.2} & \underline{64.3} & \underline{51.8} & \better{19.8\%$\uparrow$} & \underline{0.299} & \better{4.1\%$\downarrow$} \\
Ouro & 13.5 & 11.2 & 45.6 & 40.4 & 75.4 & 56.9 & 52.8 & 42.3 & \worse{2.2\%$\downarrow$} & 0.321 & \worse{3.1\%$\uparrow$} \\
Hyperloop & 18.5 & 15.8 & 54.7 & 46.7 & 79.6 & 63.4 & 63.6 & 48.9 & \better{13.2\%$\uparrow$} & 0.303 & \better{2.8\%$\downarrow$} \\
TaH2-fixed & 18.8 & 12.7 & 52.2 & 43.1 & 77.0 & 64.0 & 59.8 & 46.8 & \better{8.3\%$\uparrow$} & 0.303 & \better{2.6\%$\downarrow$} \\
WMRE & 12.7 & 11.2 & 49.5 & 41.9 & 73.4 & 60.1 & 55.2 & 43.4 & \better{0.5\%$\uparrow$} & 0.312 & \worse{0.1\%$\uparrow$} \\
CloseLoop & 18.3 & 13.8 & 55.5 & 44.1 & 77.0 & 64.8 & 58.3 & 47.4 & \better{9.7\%$\uparrow$} & 0.307 & \better{1.4\%$\downarrow$} \\
InfiLoop & 13.5 & 10.8 & 49.8 & 40.9 & 76.4 & 57.9 & 55.8 & 43.6 & \better{0.9\%$\uparrow$} & 0.312 & \worse{0.0\%$\uparrow$} \\
\padbelow BRT & 21.2 & 13.8 & 54.7 & 44.7 & 77.0 & 64.2 & 59.4 & 47.9 & \better{10.7\%$\uparrow$} & 0.305 & \better{2.0\%$\downarrow$} \\
\shlineshaded
\padabove\padbelow\textbf{\method{}} & \textbf{30.2} & \textbf{24.6} & \textbf{67.5} & \textbf{52.9} & \textbf{84.2} & \textbf{75.3} & \textbf{69.8} & \textbf{57.8} & \better{\textbf{33.7}\%$\uparrow$} & \textbf{0.291} & \better{\textbf{6.8}\%$\downarrow$} \\
\noalign{{\color{baselinecolor}\hrule height 1pt}}
\end{tabular}
\end{table*}

\section{Experiments}
\label{sec:exp}

\begin{figure}[t]
  \vspace{4pt}
  \centering
  \includegraphics[width=\columnwidth]{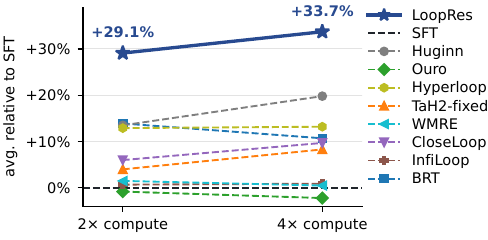}
  \vspace{-15pt}
  \caption{\textbf{Gain over SFT as compute doubles.}
    Each line connects a design's change in the average of the seven math and code scores at 2$\times$ and 4$\times$ compute, on Qwen3-1.7B-Base.}
  \label{fig:compute_gain}
  \vspace{-5pt}
\end{figure}

We validate the performance of \method{} and analyze its design through extensive experiments.
In particular, we investigate the following questions:
\begin{itemize}[leftmargin=*,nosep]
\item Can \method{} consistently outperform existing looped designs? (\cref{tab:main}, \cref{fig:compute_gain})
\item Does \method{} scale to deeper loops, larger backbones, and other model families? (\cref{fig:tts_loops}, \cref{tab:scale}, \cref{tab:llama})
\item How much does each component contribute? (\cref{tab:ablation})
\item What patterns does \method{} learn? (\cref{fig:block1_row}, \cref{fig:block2_final})
\end{itemize}

\subsection{Setups}

\paragraph{Models and baselines.}
We use Qwen3-\{1.7B, 4B, 8B, 30B-A3B\}-Base~\cite{yang2025qwen3}, Qwen3.5-2B-Base~\cite{qwen2026qwen35}, and Llama-3.2-1B-Base~\cite{grattafiori2024llama3} as backbones, with Qwen3-1.7B-Base unless otherwise stated.
We compare \method{} with SFT, the backbone finetuned without looping, and with 8 looped baselines.
5 of them loop all layers: Ouro~\cite{zhu2025ouro}, TaH2-fixed~\cite{you2026tah2}, CloseLoop~\cite{marchenko2026closing}, InfiLoop~\cite{li2026infiloop}, and BRT~\cite{marques2026brt}.
The other 3 loop the middle half of the layers as \method{} does: Huginn~\cite{geiping2025huginn}, Hyperloop~\cite{zeitoun2026hyperloop}, and WMRE~\cite{zhuang2026recurrence}.
All baselines are compared at 2$\times$ and 4$\times$ compute, corresponding to $L=2, 4$ for all-layer baselines and $L=3, 7$ for half-layer ones.
\method{} uses $W=1$ at both compute budgets.

\paragraph{Training setup.}
We finetune on math and code questions from AM-Qwen3-Distilled~\cite{tian2025correct} and UltraData-SFT-2605~\cite{openbmb2026ultradata}.
Qwen3-1.7B-Base trains on replies regenerated by Qwen3-8B, totalling 572K samples and 3.1B tokens; all other backbones train on the replies generated by Qwen3-235B-A22B.
All models share the same hyperparameters: 12,000 steps with a learning rate of $3\times10^{-5}$, as detailed in \cref{app:training}.

\paragraph{Evaluation setup.}
We report the loss on 1,000 evaluation samples and evaluate on seven benchmarks for backbones below 4B: MATH500~\cite{lightman2024verify}, OlympiadBench~\cite{he2024olympiadbench}, AMC23, AIME25, and AIME26 in math, and HumanEval+ and MBPP+~\cite{liu2023evalplus} in code.
Since the 4B and larger backbones saturate the easier benchmarks, we evaluate them on five harder ones instead: AIME25, AIME26, OlympiadBench, and IMO-AnswerBench~\cite{luong2025imobench} in math, and LiveCodeBench v6~\cite{jain2025livecodebench} in code.
We sample with temperature 0.6, top-p 0.95, and top-k 20 up to 32K tokens, and report pass@1 averaged over approximately 500 samples per benchmark.

\begin{table}[t]
\caption{\textbf{Training overhead.} We report the training step time and the peak activation memory of \method{} on Qwen3-1.7B-Base, measured at 16k sequence length.}
\label{tab:overhead}
\vspace{-7pt}
\newlength{\ohpct}
\newcommand{\nopct}[1]{\settowidth{\ohpct}{#1}\makebox[\ohpct][c]{--}}
\tablestyle{5.5pt}{1.1}
\begin{tabular}{l|r|r}
\padabove\padbelow variant & \multicolumn{1}{c|}{step time (ms)} & \multicolumn{1}{c}{peak mem. (GiB)} \\
\shline
\padabove\padbelow plain residual & 1886.3\quad\nopct{+54.9\%} & 8.56\quad\nopct{+47.5\%} \\
\shlineshaded
\padabove\padbelow\method{} & 1976.6\quad\phantom{0}+4.8\% & 8.62\quad\phantom{0}+0.7\% \\
\shlinefromshaded
\padabove\hspace{3pt}w/o loop-aware key & 1968.9\quad\phantom{0}+4.4\% & 8.62\quad\phantom{0}+0.7\% \\
\padbelow\hspace{3pt}w/o fused kernel & 2922.3\quad+54.9\% & 12.63\quad+47.5\% \\
\end{tabular}
\vspace{-5pt}
\end{table}

\subsection{Main Results}

\paragraph{Performance results.}
\cref{tab:main} compares \method{}, SFT, and 8 baselines across 7 benchmarks and the loss at 2$\times$ and 4$\times$ compute on Qwen3-1.7B-Base.
To ensure a fair comparison, we reimplement every baseline on the same backbone and finetune it on the same data with the same hyperparameters, as detailed in \cref{app:baselines}.

Based on the results, we make the following observations:
(i) \method{} achieves the best score on every benchmark at both compute budgets, more than doubling the gain of the strongest baseline at 2$\times$ compute.
(ii) The relative gains are largest on the hardest benchmarks (e.g., AIME), likely because harder questions benefit most from the extra computation per token that deeper loops provide.

\begin{figure*}[t]
\begin{minipage}[t]{0.32\textwidth}
\vspace{0pt}
\captionof{table}{\textbf{Generalizing across model families.} We compare backbones by average score at 2$\times$ compute.}
\label{tab:llama}
\vspace{-7pt}
\tablestyle{4pt}{1.1}
\newcommand{\llamatabular}{\begin{tabular}{l|c>{\columncolor{baselinecolor}}cc}
\padabove\padbelow backbone & SFT & \cellcolor{white}\textbf{\method{}} & $\Delta$ \\
\noalign{\vspace{1pt}}
\hline
\gaprow{\multicolumn{1}{c}{} & & & }
\padabove Llama-3.2-1B & 14.3 & \textbf{19.6} & \better{+5.4} \\
\padbelow Qwen3.5-2B & 52.1 & \textbf{64.1} & \better{+12.0} \\
\gaprow{\multicolumn{1}{c}{} & & & }
\end{tabular}}
\loop\sbox0{\llamatabular}\ifdim\dimexpr\columnwidth-\wd0\relax>0.01pt
  \setlength{\tabcolsep}{\dimexpr\tabcolsep+(\columnwidth-\wd0)/8\relax}\repeat
\llamatabular

\vspace{12.47pt}
\captionof{table}{\textbf{Token efficiency.} Average output tokens per question and per correct answer on Qwen3-1.7B-Base at 2$\times$ compute.}
\label{tab:length}
\vspace{-7pt}
\tablestyle{4pt}{1.1}
\newcommand{\lengthtabular}{\begin{tabular}{l|c>{\columncolor{baselinecolor}}cc}
\padabove\padbelow questions & SFT & \cellcolor{white}\textbf{\method{}} & $\Delta$ \\
\noalign{\vspace{1pt}}
\hline
\gaprow{\multicolumn{1}{c}{} & & & }
\padabove all & 18,019 & \textbf{15,302} & \better{15.1\%$\downarrow$} \\
\padbelow correct & 39,652 & \textbf{25,211} & \better{36.4\%$\downarrow$} \\
\gaprow{\multicolumn{1}{c}{} & & & }
\end{tabular}}
\loop\sbox0{\lengthtabular}\ifdim\dimexpr\columnwidth-\wd0\relax>0.01pt
  \setlength{\tabcolsep}{\dimexpr\tabcolsep+(\columnwidth-\wd0)/8\relax}\repeat
\lengthtabular

\end{minipage}\hfill
\begin{minipage}[t]{0.375\textwidth}
\vspace{0pt}
\captionof{table}{\textbf{Ablation study.} We vary the components of \method{} on Qwen3-1.7B-Base (default in gray).}
\label{tab:ablation}
\vspace{-7pt}
\tablestyle{4pt}{1.1}
\newcommand{\default}[1]{{\setlength{\fboxsep}{1.5pt}\colorbox{baselinecolor}{#1}}}
\newcommand{\group}[2]{\multicolumn{5}{l}{\padabove\emph{#1} (\method{}: \default{#2})}}
\newcommand{\variant}[1]{\hspace{3pt}#1}
\newcommand{\up}[1]{\better{+#1}}
\newcommand{\down}[1]{\worse{$-$#1}}
\newcommand{\ablationtabular}{\begin{tabular}{l|cc|cc}
\padbelow\headspan{variant} & \multicolumn{2}{c|}{2$\times$ compute} & \multicolumn{2}{c}{4$\times$ compute} \\
\cmidrule(lr){2-3}\cmidrule(lr){4-5}
\padabove\padbelow& avg. & $\Delta$ & avg. & $\Delta$ \\
\shlineshaded
\padabove\padbelow\method{} & 55.8 & -- & 57.8 & -- \\
\shlinefromshaded
\padabove\padbelow Huginn + AttnRes & 53.9 & \down{1.9} & 55.0 & \down{2.8} \\
\shline
\group{Residual window}{$W=1$} \\
\variant{$W=2$} & 55.2 & \down{0.6} & 57.5 & \down{0.3} \\
\padbelow\variant{$W=3$} & -- & -- & 57.7 & \down{0.1} \\
\shline
\group{Key}{loop-aware + content} \\
\variant{loop-aware only} & 54.9 & \down{0.9} & 57.5 & \down{0.3} \\
\padbelow\variant{content only} & 54.9 & \down{0.9} & 55.8 & \down{2.0} \\
\gaprow{\multicolumn{1}{c}{} & & & & }
\end{tabular}}
\loop\sbox0{\ablationtabular}\ifdim\dimexpr\columnwidth-\wd0\relax>0.01pt
  \setlength{\tabcolsep}{\dimexpr\tabcolsep+(\columnwidth-\wd0)/10\relax}\repeat
\ablationtabular
\end{minipage}%
\hfill
\input{figures_tex/tts_loops}
\end{figure*}

\begin{table}[t]
\caption{\textbf{Scaling to larger backbones.} We compare \method{} with SFT at 2$\times$ compute on AIME25~(A25), AIME26~(A26), OlympiadBench~(Oly), IMO-AnswerBench~(IMO), and LiveCodeBench v6~(LCB) for larger backbones.}
\label{tab:scale}
\newcommand{\up}[1]{\better{+#1}}
\newcommand{\band}{\rule{0pt}{\dimexpr\ht\strutbox+1pt\relax}}
\newcommand{\group}[1]{\multicolumn{8}{l}{\padabove\emph{#1}}}
\newcommand{\bandend}{\noalign{{\color{baselinecolor}\hrule height 1pt}}}
\tablestyle{4pt}{1.1}
\newcommand{\scaletabular}{\begin{tabular}{l|ccccc|cc}
\padbelow method & A25 & A26 & Oly & IMO & LCB & avg. & $\Delta$ \\
\shline
\group{Qwen3-4B-Base} \\
\hspace{3pt}SFT & 35.4 & 38.3 & 61.6 & 20.5 & 27.4 & 36.6 & -- \\
\rowcolor{baselinecolor}\band\hspace{3pt}\textbf{\method{}} & \textbf{47.1} & \textbf{51.0} & \textbf{68.3} & \textbf{29.6} & \textbf{32.9} & \textbf{45.8} & \up{9.2} \\
\shlinefromshaded
\group{Qwen3-8B-Base} \\
\hspace{3pt}SFT & 39.6 & 46.0 & 65.5 & 30.9 & 32.6 & 42.9 & -- \\
\rowcolor{baselinecolor}\band\hspace{3pt}\textbf{\method{}} & \textbf{51.5} & \textbf{62.9} & \textbf{72.3} & \textbf{35.5} & \textbf{35.1} & \textbf{51.5} & \up{8.6} \\
\shlinefromshaded
\group{Qwen3-30B-A3B-Base} \\
\hspace{3pt}SFT & 59.6 & 66.3 & 72.9 & 38.6 & 38.3 & 55.1 & -- \\
\rowcolor{baselinecolor}\band\hspace{3pt}\textbf{\method{}} & \textbf{65.4} & \textbf{72.5} & \textbf{76.4} & \textbf{44.9} & \textbf{44.0} & \textbf{60.6} & \up{5.5} \\
\bandend
\end{tabular}}
\loop\sbox0{\scaletabular}\ifdim\dimexpr\columnwidth-\wd0\relax>0.01pt
  \setlength{\tabcolsep}{\dimexpr\tabcolsep+(\columnwidth-\wd0)/16\relax}\repeat
\scaletabular
\vspace{-5pt}
\end{table}

\paragraph{Loop gains.}
\cref{fig:compute_gain} illustrates how the gain over SFT grows as compute doubles.
Most baselines gain little or even regress, and the few that continue to improve remain well below \method{}.
In contrast, \method{} converts the extra loops into further gains from an already high score, showing that its advantage holds as loops deepen.

\paragraph{Token efficiency.}
\cref{tab:length} compares the token cost of \method{} with SFT at 2$\times$ compute.
\method{} spends fewer tokens than SFT over all questions and further reduces the effort per correct answer, suggesting that the loops make the model more capable and answer in fewer tokens.

\paragraph{Overhead.}
\cref{tab:overhead} compares the training cost of \method{} with a plain residual baseline.
\method{} adds little computational overhead and almost no activation memory due to its residual window and fused kernel, and the loop-aware key itself is nearly free.
The naive PyTorch implementation is far more expensive even with the window.

\subsection{Scalability}\label{sec:generalization}

\paragraph{Scaling to deeper loops.}
We train \method{} with different numbers of loops on Qwen3-1.7B-Base and evaluate each across thinking budgets.
As shown in \cref{fig:tts_loops}, the score improves as loops deepen and gradually saturates beyond $L=5$ (3$\times$ compute).
The best trade-off between score and compute is at $L=3$ (2$\times$ compute).

\paragraph{Scaling to larger backbones.}
We apply \method{} to Qwen3-\{4B,8B\}-Base and the MoE Qwen3-30B-A3B-Base at 2$\times$ compute.
As shown in \cref{tab:scale}, \method{} improves every benchmark on all three backbones.
\cref{fig:teaser} and \panelref{fig:scale_backbones}{a} further show that its score rises with a clearly steeper slope than SFT as decode FLOPs grow, especially once reasoning begins to saturate.
This advantage is more evident on smaller models.
Notably, \method{} on the 4B backbone even surpasses SFT on the 8B one in average score.

\paragraph{Generalizing across model families.}
We further apply \method{} to Llama-3.2-1B-Base and Qwen3.5-2B-Base, whose architectures and behaviors differ from Qwen3.
As shown in \cref{tab:llama} and \cref{tab:app_llama}, \method{} improves scores on both families.
Across all backbones, \method{} consistently delivers gains, demonstrating its generalizability.

\subsection{Ablation Study}
\label{sec:ablation}

We ablate the components of \method{} on Qwen3-1.7B-Base at both compute budgets in \cref{tab:ablation}.

\paragraph{Naive AttnRes.}
We first apply AttnRes to the core of Huginn without loop-aware keys or the residual window as in \cref{sec:attnres_loop}.
This naive combination performs clearly worse than \method{} at both compute budgets, with the gap widening as loops deepen.

\begin{figure*}[t]
  \centering
  \includegraphics[width=\textwidth]{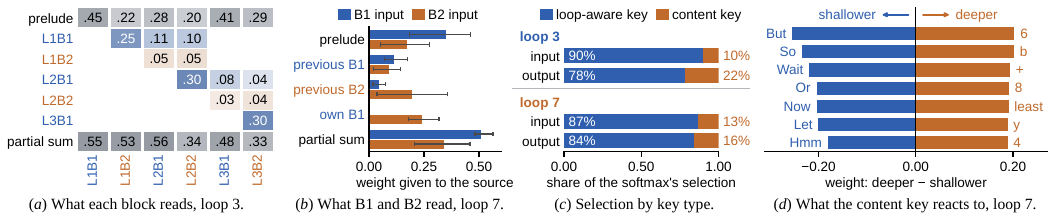}
  \vspace{-10pt}
  \caption{\textbf{What \method{} learns.}
    \method{} on Qwen3-1.7B-Base, averaged over 128 samples. L$t$B$j$ denotes the output of block $j$ in loop $t$.
    (a) Source weights for each block at $L{=}3$.
    (b) Source weights of B1 and B2 at $L{=}7$, averaged over loops with min--max bars.
    (c) Contribution of the loop-aware key and the content key to source selection, measured by Shapley values.
    (d) Tokens ranked by whether their content key steers B2 toward its shallower or deeper output at $L{=}7$.}
  \label{fig:block1_row}
\end{figure*}

\begin{figure*}[t]
  \centering
  \includegraphics[width=\textwidth]{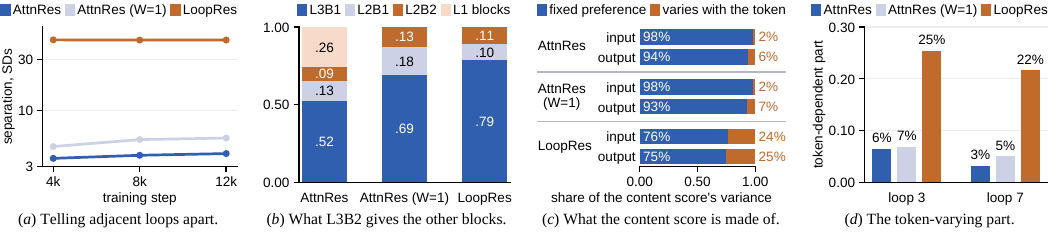}
  \vspace{-10pt}
  \caption{\textbf{How \method{} differs from AttnRes.}
    AttnRes, and AttnRes with $W{=}1$ and \method{} on Qwen3-1.7B-Base over 128 samples at $L{=}3$.
    (a) Output difference between adjacent loops for each block, measured in standard deviations across tokens.
    (b) Source weights of L3B2 over earlier blocks, renormalized to exclude the prelude and partial sum.
    (c) Content-score variance across sources, decomposed into a fixed-preference component and a token-varying component.
    (d) Token-varying fraction of the content-score variance at loop outputs, for $L{=}3$ and $L{=}7$.}
  \label{fig:block2_final}
\end{figure*}

\paragraph{Residual window.}
We vary the window size $W$, which bounds how many previous loops each layer takes as input.
A larger window reduces the average score at both compute budgets.
Thus, we set $W=1$, as it achieves the best score at the lowest cost.
However, adding this window alone to naive AttnRes (i.e., the content-only variant) brings only a slight improvement and remains worse than \method{}.

\paragraph{Loop-aware keys.}
We then examine the contribution of the loop-aware and content keys.
Either key alone performs worse than the two combined at both compute budgets.
As loops deepen, dropping the loop-aware key hurts more, while using it alone approaches \method{}.
We hypothesize that as loops deepen, iterates become harder to distinguish by content, so the model relies more on the structural information carried by the loop-aware key.

\subsection{Analysis}
\label{sec:analysis}

\paragraph{What \method{} learns.}
We visualize how \method{} weighs its sources, revealing three patterns.
\begin{itemize}[leftmargin=*,nosep]
\item \textbf{Iterative loop updates.} In \panelref{fig:block1_row}{a,b}, each layer balances its input across the prelude, the loop outputs, and the partial sum.
Within the loop outputs, the first block mainly updates its own output from the previous loop, with a small share from the other block of that loop.
The second block on average relies most on the first block of the current loop, indicating that \method{} learns the iterative structure of the loop.
\item \textbf{Dominant loop-aware keys.} In \panelref{fig:block1_row}{c}, the loop-aware key accounts for most of the selection at both compute budgets, with the content key taking a smaller share.
\item \textbf{Adaptive content keys.} Despite its smaller share, the content key lets each token choose its own depth.
In \panelref{fig:block1_row}{d}, simple tokens such as ``But'' and ``Wait'' favor shallower outputs, while difficult tokens such as digits and math symbols favor deeper ones.
This suggests that the content key recognizes the difficulty of each token and allocates computation accordingly.
\end{itemize}

\paragraph{How \method{} differs from AttnRes.}
We further compare \method{} with AttnRes, with and without the residual window, revealing three differences.
\begin{itemize}[leftmargin=*,nosep]
\item \textbf{Distinguishable loop iterates.} In \panelref{fig:block2_final}{a}, AttnRes struggles to tell adjacent loops apart even with the residual window, while \method{} separates them clearly in the early stage of finetuning.
\item \textbf{Recency-focused inputs.} In \panelref{fig:block2_final}{b}, the last block of AttnRes gives a sizable share of its weight to the first loop.
The window removes these shortcuts, and \method{} further concentrates the weight on the preceding block.
\item \textbf{Less is more for content keys.} In \panelref{fig:block2_final}{c,d}, the content score of AttnRes is almost a fixed preference over sources and barely varies with the token.
Although the content key takes a smaller share in \method{}, its score varies far more with the token, even as loops deepen, likely because the loop-aware key takes over the fixed preference.
\end{itemize}

\section{Related Work}
\label{sec:related}

\paragraph{Looped Transformers.}
Looped Transformers reuse a shared block of layers multiple times in a single forward pass, decoupling depth from the number of parameters.
Early work and theoretical studies show that this recurrence is expressive, letting a fixed set of layers carry out computation that would otherwise require a deeper model~\cite{dehghani2019universal,giannou2023looped,yang2024looped,fan2025looped,saunshi2025reasoning}.
Recent work brings loops to LLMs, studying how to pretrain them at scale~\cite{geiping2025huginn,zhu2025ouro,gao2026loopie,fishman2026dory}, how to pair loops with conditional compute~\cite{csordas2024moeut,bae2025mor}, how loops trade off against parameters and data~\cite{prairie2026parcae,wang2026smelt,ekbote2026loop,huang2026looped}, how to convert pretrained LLMs into looped ones~\cite{bae2025relaxed,koishekenov2025etd,mcleish2025retrofitted,marchenko2026closing}, and how to train one set of weights that serves every loop~\cite{prairie2026parcae,zhu2026loopdropout,jeddi2026loopformer}.
Orthogonal to these questions, we next review looped designs by how the outputs of earlier loops reach later loops.

\paragraph{Passing one state across loops.}
Most looped designs treat a loop as a recurrence over a single hidden state, in which each loop takes the output of the previous loop as its residual input~\cite{lan2020albert,xue2022go,takase2023lessons,zhu2025ouro,bae2025relaxed,koishekenov2025etd,marques2026brt,gao2026loopie,fu2026tah}.
Many further inject the input embedding into every loop, keeping later loops conditioned on the input as the state is refined~\cite{geiping2025huginn,mcleish2025retrofitted,prairie2026parcae,marchenko2026closing,you2026tah2,fishman2026dory,huang2026looped}.
However, the previous state is the only channel through which a loop sees earlier loops, so everything earlier loops compute must survive in this one state, which limits what later loops can build on.

\paragraph{Aggregating states across loops.}
To let later loops draw on more than the latest state, another line of work carries information from several earlier loops into the next one.
One branch keeps a fixed-size memory of earlier states and merges it into the residual input of each loop~\cite{zeitoun2026hyperloop,zhuang2026recurrence,li2026infiloop}, and another lets attention reuse the keys and values computed in earlier loops~\cite{bae2025mor,fu2026tah,zhu2026mixtureofloopstransformer}.
Nevertheless, the residual input in both branches still compresses the outputs of earlier loops into a narrow pathway.
\method{} instead keeps each block output as a separate source and lets every sub-layer weigh them.

\paragraph{Residual connections.}
The residual stream is the pathway through which each layer reads the outputs of preceding layers.
Residual connections~\cite{he2016deep} sum these outputs with fixed unit weights and remain standard in modern LLMs, but the contribution of each layer dilutes as depth grows~\cite{sun2025curse,zhu2026mixtureofdepthsattention}.
Several works widen this pathway so that each layer draws on preceding layers more directly~\cite{huang2017densely,pagliardini2024denseformer,zhu2025hyperconnections,xiao2025muddformer,xie2025mhc,zhu2026mixtureofdepthsattention}, and AttnRes~\cite{kimi2026attnres} lets each layer attend to the outputs of preceding layers with learned, input-dependent weights.
However, these designs target non-looped networks, where the sources differ in content and are fixed in number.
\method{} builds on AttnRes and adapts it to loops, where the sources are iterates of the same layers and grow with the loop count.
It lets every layer build its input by attending directly to the sources of the latest $W$ loops, with a loop-aware key that tells the iterates apart.
\section{Conclusion}
\label{sec:concl}

In this work, we presented \method{}, an effective, scalable, and simple approach that lets each looped layer attend directly to the outputs of earlier loops.
\method{} assigns each source a trainable loop-aware key that encodes the structural relationship of loops and layers, and restricts each layer to a residual window of the latest $W$ loops so that every loop builds incrementally on recent ones.
Across 6 pretrained LLMs from 3 model families and 9 math and code benchmarks, \method{} delivered significant and consistent gains, more than doubling the improvement of the strongest baseline at 2$\times$ compute.
Its score rose with a steeper slope than SFT as compute grew, reaching a higher ceiling and needing less compute to match SFT on long outputs.
These advantages benefit both inference and post-training rollouts.
We hope \method{} serves as a simple and effective foundation for scaling LLMs along the loop-depth axis.

\paragraph{Limitations.}
Our experiments cover math and code benchmarks, and the gains on broader tasks such as general instruction following and open-ended generation remain to be validated.
Additionally, \method{} has been evaluated with supervised finetuning only, and its interaction with reinforcement learning has not yet been explored.

\paragraph{Future work.}
One key direction is to combine \method{} with reinforcement learning, where the lower cost of matching SFT on long outputs could reduce the rollout budget.
Another direction is to explore adaptive loop depth, allowing the model to decide how many loops each token or sequence needs.
Furthermore, our findings suggest that the loop-depth scaling curve has not yet saturated on larger backbones, motivating further study of looping at greater scale.
Finally, our analysis reveals that the learned loop dependencies follow clear, structured patterns, suggesting that a simpler, fixed connection scheme may capture most of the benefit without the overhead of learned attention weights.

\ifarxiv
  \section*{Acknowledgements}

We thank Fuzhao Xue, Tong Li, and Yong Liu for helpful discussions and feedback on this work. Yang You's research group is being sponsored by ByteDance grant, Alibaba grant, Meta Research Award, AliCloud grant, Apple grant, Google Research Award, Microsoft Research grant, and Adobe gift. We are grateful to NUSIT’s Research Computing group for the computing support with numbers CFP03-CF-088 and CFP05-CF-062.

\fi
\section*{Impact Statement}

\method{} improves the math and code performance of pretrained LLMs by finetuning them into loops, rather than pretraining larger models from scratch.
This may lower the cost of obtaining capable models and broaden access for researchers and practitioners.
However, stronger reasoning and coding capabilities may also be misused, for example to generate malicious code.
We therefore encourage safety evaluations and safeguards before deployment.
Finally, looping increases the computation per token, which may raise energy consumption at scale.

\bibliography{references}
\bibliographystyle{icml2026}

\newpage
\appendix
\onecolumn
\raggedbottom
\setlength{\intextsep}{4pt plus 2pt minus 1pt}
\vspace*{16.5pt}
\noindent\begin{minipage}{\textwidth}
  \setlength{\parskip}{6pt}
  {\centering
    {\Large\bfseries Appendix\par}}
  \vskip 9pt

  We organize our appendix as follows:

  \textbf{Experimental Details:}
  \begin{itemize}
    \item Section~\ref{app:config}: Model configurations
    \item Section~\ref{app:baselines}: Baselines
    \item Section~\ref{app:data}: Data
    \item Section~\ref{app:training}: Training
    \item Section~\ref{app:eval}: Evaluation
    \item Section~\ref{app:pseudo}: Pseudo code
  \end{itemize}

  \textbf{Supplementary Results:}
  \begin{itemize}
    \item Section~\ref{app:bench}: Detailed benchmark scores
    \item Section~\ref{app:loss}: Validation loss over finetuning
    \item Section~\ref{app:tts}: Test-time scaling on more backbones
    \item Section~\ref{app:tokens}: Performance under equal token budgets
  \end{itemize}

  \textbf{Additional Experiments:}
  \begin{itemize}
    \item Section~\ref{app:huginn_all}: Alternative ways of looping with AttnRes
    \item Section~\ref{app:attnres_noloop}: Applying AttnRes without looping
    \item Section~\ref{app:data2x}: Comparison with SFT at matched training compute
    \item Section~\ref{app:4b8b}: Looping a 4B model against a non-looped 8B model
  \end{itemize}
\end{minipage}

\section{Experimental Details}
\label{app:setup}

\subsection{Model Configurations}
\label{app:config}
\begin{table}[H]
  \caption{\textbf{Configuration of \method{} on each backbone.}
    $L$ is given at $2\times$ / $4\times$ compute, and -- marks a budget not run.}
  \label{tab:app_config}
  \tablestyle{8pt}{1.15}
  \begin{tabular}{lccccc}
    \toprule
    \padabove\padbelow Backbone & Layers & Prelude / Core / Coda & $B$ & $L$ & $W$ \\
    \midrule
    \padabove Qwen3-1.7B-Base & 28 & 7 / 14 / 7  & 2 & 3 / 7  & 1 \\
    Qwen3-4B-Base      & 36 & 9 / 18 / 9  & 2 & 3 / -- & 1 \\
    Qwen3-8B-Base      & 36 & 9 / 18 / 9  & 2 & 3 / -- & 1 \\
    Qwen3-30B-A3B-Base & 48 & 12 / 24 / 12 & 2 & 3 / -- & 1 \\
    Qwen3.5-2B-Base    & 24 & 6 / 12 / 6  & 2 & 3 / -- & 1 \\
    Llama-3.2-1B-Base       & 16 & 4 / 8 / 4   & 2 & 3 / -- & 1 \\
  \end{tabular}
\end{table}

\cref{tab:app_config} lists the configuration of \method{} on each backbone.
We split each backbone into a prelude, a core, and a coda of one quarter, one half, and one quarter of its layers, and cut the core into $B=2$ blocks.
We measure compute by the number of layers each token passes through relative to the backbone, so running the core $L=3$ and $L=7$ times costs 2 and 4 full-model passes, respectively.

\subsection{Baselines}
\label{app:baselines}
We reimplement every baseline on the same backbone and train it with its own objective, the cross-entropy of the last loop unless stated otherwise.
Since two of them are unnamed, we call them WMRE~\cite{zhuang2026recurrence} and CloseLoop~\cite{marchenko2026closing} after their titles.
The originals of Ouro, InfiLoop, BRT, Huginn, Hyperloop, and WMRE pretrain from scratch; we finetune the pretrained backbone throughout.
Every token runs all $L$ loops, in training and at inference, and the three core-looped designs run the coda once, on the last loop's output.

\paragraph{Ouro~\cite{zhu2025ouro}.}
Our implementation loops all layers, each loop reading the previous loop's normed output, and averages the cross-entropy over all loops.
The original weights the loops by a learned exit distribution under an entropy-regularized objective; we remove the exit gate and weight the loops uniformly.

\paragraph{TaH2-fixed~\cite{you2026tah2}.}
The paper's fixed-depth variant, following its released recipe.
It keeps TaH2's updater, a SwiGLU block as wide as the model that merges the token embeddings with the previous loop's output, and removes the iteration decider.
Attention runs in the released code's causal mode: earlier tokens expose their loop-1 keys, and each token reads its own keys of the later loops.
The model predicts from the mean of the loops' next-token distributions, and the objective is the cross-entropy of this mean.

\paragraph{CloseLoop~\cite{marchenko2026closing}.}
Loop $t>1$ reads $w\ve + (1-w)\vh_{t-1}$, with $w$ the sigmoid of one trained scalar at the paper's $w=0.99$, and the LM head is untied, starting as a copy of the input embeddings.
An exit gate scores each loop, and the objective weights the loops' cross-entropy by the exit distribution, with the paper's smoothing 0.05 and entropy coefficient 0.15.
We omit the paper's optional loop-specific LoRA adapters and always predict from loop $L$.

\paragraph{InfiLoop~\cite{li2026infiloop}.}
Each loop reads a running average of the earlier loops' outputs, weighted by a trained decay and a content score, $\exp(\vw^\top \mathrm{RMSNorm}(\vz_t))$ clamped to $[-8, 8]$, with the released code's initialization ($\vw$ at zero, the decay at 0.55).
The original runs a two-level recurrence with one average per level, restarts the inner one each cycle, seeds the outer one with a learned state, and backpropagates only through the last cycle; we keep a single average over the $L$ loops, seed it with the token embeddings, and backpropagate through all of them.
Its post-norm layers also keep the averaged states at unit scale, while we average the raw residual stream of the pre-norm backbone.

\paragraph{BRT~\cite{marques2026brt}.}
Consecutive layers are grouped into blocks, and each block runs $L$ times on the standard residual stream before the next.
We use seven blocks of four layers on the 28-layer backbone, where the original defaults to two blocks.

\paragraph{Huginn~\cite{geiping2025huginn}.}
Retrofitted with the same prelude, core, and coda as \method{}: in each loop the core reads Parcae's diagonal injection~\cite{prairie2026parcae} of the previous core output and the prelude output, initialized at its retention $\sqrt{1/5}$ with the injection the identity, so that one loop reproduces the pretrained model.
The original samples the loop count from a log-normal Poisson distribution, draws the initial state from a truncated normal, and joins the two inputs by a linear map of their concatenation; we fix the loop count, start the state at zero, and use the diagonal adapter.

\paragraph{Hyperloop~\cite{zeitoun2026hyperloop}.}
The core runs on four parallel residual streams, joined after each loop by its own sigmoid-gated hyper-connection, with a per-loop vector trained from zero added to the core output, and the coda reads the stream mean.
The gate scales start at mHC's 0.01~\cite{xie2025mhc}, and the biases start the $\mathrm{pre}$, $\mathrm{post}$, and $\mathrm{res}$ gates at $1/4$, $1$, and $1/2$.
The stream count and the begin--middle--end split match the original.

\paragraph{WMRE~\cite{zhuang2026recurrence}.}
The paper's H+LG configuration, on the same split as \method{}: loop $t+1$ reads $\vh_t + \sum_{j} \vb_j \odot (\vh_{t-j} - \vh_t)$ over the lags $j \le 2$ that reach back to the first loop, with $\vh_t$ the state after loop $t$, and a scalar gate over the paper's eight time features scales each loop's update.
Both sets of weights train from zero.

\subsection{Data}
\label{app:data}
We take the math and code questions of AM-Qwen3-Distilled~\cite{tian2025correct} and UltraData-SFT-2605~\cite{openbmb2026ultradata} and answer each with Qwen3-8B in thinking mode, sampling with temperature 0.6, top-p 0.95, and top-k 20 up to 16,384 tokens.
A judge model Qwen3-235B-A22B checks whether each reply reaches the same final answer as the original reply, and we keep only the consistent replies.
We further drop malformed replies and every question that shares a run of 13 words with evaluation benchmarks.
Every other backbone trains on the same questions answered by Qwen3-235B-A22B.

\begin{figure}[!t]
\begin{lstlisting}[style=pycode, linebackgroundcolor={%
  \color{white}%
  \ifnum\value{lstnumber}=14\color{codehl}\fi
  \ifnum\value{lstnumber}>22\ifnum\value{lstnumber}<25\color{codehl}\fi\fi
  \ifnum\value{lstnumber}>33\ifnum\value{lstnumber}<39\color{codehl}\fi\fi}]
def loop_res(values: list[Tensor], loop_keys: list[Tensor], loop_query: Tensor, content_query: Tensor,
             norm: RMSNorm) -> Tensor:
    """
    AttnRes input with loop-aware keys.
    values:
        S sources of shape [N, T, D]
    loop_keys:
        S loop-aware keys of shape [D], one trainable vector per source
    """
    values = torch.stack(values)        # [S, N, T, D]
    loop_keys = torch.stack(loop_keys)  # [S, D]
    content_keys = norm(values)         # [S, N, T, D], as in AttnRes
    logits = torch.einsum('d, s n t d -> s n t', content_query, content_keys)
    logits = logits + torch.einsum('d, s d -> s', loop_query, loop_keys)[:, None, None]  # loop-aware keys
    weights = logits.softmax(0)
    return torch.einsum('s n t, s n t d -> n t d', weights, values)

def forward(self, prelude_out: Tensor) -> Tensor:
    """
    prelude_out:
        [N, T, D]: output of the prelude; the returned core output goes to the coda
    """
    # residual window: the blocks of the previous W loops and their loop-aware keys
    window_values, window_keys = [], []
    for t in range(self.num_loops):
        values = [prelude_out] + window_values
        loop_keys = [self.prelude_key] + window_keys

        # block AttnRes over the core; every sub-layer forms its input by loop_res, with its own queries in loop t
        blocks = []
        for i, block in enumerate(self.core_blocks):
            blocks.append(block(values + blocks, loop_keys + self.block_keys[t][:i], loop=t))

        # end of loop t: once the window is full, drop all blocks of its oldest loop, then add loop t
        if len(window_values) == self.window_size * self.num_blocks:
            del window_values[:self.num_blocks], window_keys[:self.num_blocks]
        window_values += blocks
        window_keys += self.block_keys[t]

    # output: the window and the blocks of the last loop
    return loop_res(values + blocks, loop_keys + self.block_keys[t],
                    self.output_loop_query, self.output_content_query, self.output_norm)
\end{lstlisting}
\vspace{-4pt}
\caption{\textbf{PyTorch-style pseudo code for \method{}.} Lines in blue mark its two changes to Block AttnRes: (i) \texttt{loop\_res} adds a loop-aware key to each source. (ii) Each loop takes a residual window of the latest $W$ loops as input, from which older loops are dropped.}
\label{fig:pseudo}
\end{figure}

\subsection{Training}
\label{app:training}
\cref{tab:app_hparams} lists the hyperparameters.
Each step packs several conversations per sequence, FlashAttention-2~\cite{dao2024flashattention2} restricts attention to within each conversation, and only the reply tokens contribute to the loss.
We sweep the learning rate over $\{3\times10^{-6}, 1\times10^{-5}, 3\times10^{-5}, 5\times10^{-5}, 1\times10^{-4}\}$ and keep $3\times10^{-5}$ for every model.

\begin{table}[H]
  \caption{\textbf{Training hyperparameters}, shared by all models.}
  \label{tab:app_hparams}
  \tablestyle{8pt}{1.15}
  \begin{tabular}{ll}
    \toprule
    \padabove\padbelow Hyperparameter & Value \\
    \midrule
    \padabove Optimizer & AdamW~\cite{loshchilov2019adamw}, $\beta_1=0.9$, $\beta_2=0.95$ \\
    Weight decay & 0.01 on matrices, 0 on vectors \\
    Learning rate & $3\times10^{-5}$, linear warmup over 3\% of steps, cosine decay to $3\times10^{-6}$ \\
    Gradient clipping & 1.0 \\
    Training steps & 12,000 \\
    Batch size & 16 sequences \\
    Sequence length & 16,384 tokens \\
    Sequence packing & Random order, attention within each sample \\
    Precision & bfloat16 mixed precision \\
  \end{tabular}
\end{table}

\subsection{Evaluation}
\label{app:eval}
We run the benchmarks on the final checkpoint at step 12,000.
Replies are generated in thinking mode under the chat template, with temperature 0.6, top-p 0.95, and top-k 20 up to 32,768 new tokens.
HumanEval+ and MBPP+ run both the base and the extended tests of EvalPlus~\cite{liu2023evalplus}.
We report pass@1 averaged over the samples of each problem, and \cref{tab:app_benchmarks} lists the number of problems, the samples per problem, and the backbones evaluated on each benchmark.

For the test-time scaling curves, we truncate the thinking at each budget and force the model to start its reply, so their scores differ from those in the tables, which use the full replies.

\begin{table}[H]
\caption{\textbf{Evaluation benchmarks.} We give the problems, samples per problem, total samples, and evaluated backbones of each benchmark. We average pass@1 over the samples of each problem. HE+ denotes HumanEval+, IMO-AB IMO-AnswerBench, and LCB LiveCodeBench v6.}
\label{tab:app_benchmarks}
\vspace{-7pt}
\tablestyle{5pt}{1.2}
\begin{tabular}{l|cccccc|ccc}
\padbelow\headspan{benchmark} & \multicolumn{6}{c|}{math} & \multicolumn{3}{c}{code} \\
\cmidrule(lr){2-7}\cmidrule(lr){8-10}
\padabove\padbelow& AIME25 & AIME26 & AMC23 & Olympiad & MATH500 & IMO-AB & HE+ & MBPP+ & LCB \\
\shline
\padabove problems & 30 & 30 & 40 & 675 & 500 & 400 & 164 & 378 & 175 \\
samples per problem & 16 & 16 & 16 & 1 & 1 & 2 & 4 & 2 & 2 \\
total samples & 480 & 480 & 640 & 675 & 500 & 800 & 656 & 756 & 350 \\
\padbelow backbones & all & all & $<$4B & all & $<$4B & $\geq$4B & $<$4B & $<$4B & $\geq$4B \\
\end{tabular}
\end{table}

\subsection{Pseudo Code}
\label{app:pseudo}
\cref{fig:pseudo} provides PyTorch-style pseudo code for \method{}.

\section{Supplementary Results}
\label{app:results}

\subsection{Detailed Benchmark Scores}
\label{app:bench}
\paragraph{Ablation study.}
\cref{tab:app_ablation} reports the per-benchmark scores of the variants in \cref{tab:ablation}.

\begin{table}[H]
\caption{\textbf{Per-benchmark scores of the ablation study.} The variants are those of \cref{tab:ablation} on Qwen3-1.7B-Base, and \method{} uses $W=1$ at both compute budgets. HE+ denotes HumanEval+, and both code benchmarks report the plus tests. \emph{avg.} averages the seven math and code scores.}
\label{tab:app_ablation}
\vspace{-7pt}
\tablestyle{7.5pt}{1.2}
\begin{tabular}{l|ccccc|cc|c}
\padbelow\headspan{variant} & \multicolumn{5}{c|}{math} & \multicolumn{2}{c|}{code} & \headspan{avg.} \\
\cmidrule(lr){2-6}\cmidrule(lr){7-8}
\padabove\padbelow& AIME25 & AIME26 & AMC23 & Olympiad & MATH500 & HE+ & MBPP+ & \\
\shline
\multicolumn{9}{c}{\padabove\padbelow\textbf{2$\times$ compute}} \\
\shlineshaded
\padabove\padbelow\method{} & 27.9 & 19.4 & 64.7 & 53.8 & 84.0 & 71.8 & 68.9 & 55.8 \\
\shlinefromshaded
\padabove Huginn + AttnRes & 25.4 & 19.0 & 62.3 & 52.0 & 82.6 & 71.2 & 65.1 & 53.9 \\
window: $W=2$ & 28.1 & 18.3 & 62.2 & 53.5 & 83.0 & 72.6 & 68.9 & 55.2 \\
key: loop-aware only & 25.8 & 21.5 & 62.3 & 49.9 & 84.8 & 71.0 & 68.8 & 54.9 \\
\padbelow key: content only & 25.8 & 20.0 & 62.2 & 51.9 & 83.6 & 72.7 & 68.0 & 54.9 \\
\shline
\multicolumn{9}{c}{\padabove\padbelow\textbf{4$\times$ compute}} \\
\shlineshaded
\padabove\padbelow\method{} & 30.2 & 24.6 & 67.5 & 52.9 & 84.2 & 75.3 & 69.8 & 57.8 \\
\shlinefromshaded
\padabove Huginn + AttnRes & 27.9 & 21.0 & 63.1 & 52.4 & 83.6 & 71.2 & 66.0 & 55.0 \\
window: $W=2$ & 31.0 & 22.3 & 65.0 & 54.8 & 85.6 & 74.7 & 68.8 & 57.5 \\
window: $W=3$ & 29.0 & 24.2 & 68.0 & 52.4 & 87.2 & 74.2 & 68.8 & 57.7 \\
key: loop-aware only & 31.0 & 23.1 & 65.9 & 55.1 & 84.2 & 73.2 & 69.7 & 57.5 \\
key: content only & 27.7 & 22.9 & 63.0 & 54.8 & 83.6 & 72.6 & 66.1 & 55.8 \\
\end{tabular}
\end{table}

\paragraph{Loop depths.}
\cref{tab:app_depth} reports the per-benchmark scores of the loop depths in \cref{fig:tts_loops}.

\begin{table}[H]
\caption{\textbf{Per-benchmark scores across loop depths.} Each row runs the core $L$ times on Qwen3-1.7B-Base, at the per-token compute in parentheses relative to SFT. HE+ denotes HumanEval+, and both code benchmarks report the plus tests. \emph{avg.} averages the seven math and code scores. \textbf{Bold} marks the best score of each column.}
\label{tab:app_depth}
\vspace{-7pt}
\tablestyle{5.2pt}{1.2}
\begin{tabular}{l|ccccc|cc|c}
\padbelow\headspan{depth} & \multicolumn{5}{c|}{math} & \multicolumn{2}{c|}{code} & \headspan{avg.} \\
\cmidrule(lr){2-6}\cmidrule(lr){7-8}
\padabove\padbelow& AIME25 & AIME26 & AMC23 & Olympiad & MATH500 & HE+ & MBPP+ & \\
\shline
\padabove\padbelow SFT & 14.6 & 11.7 & 48.6 & 40.7 & 73.4 & 57.3 & 56.2 & 43.2 \\
\shline
\padabove $L=2$ (1.5$\times$) & 22.9 & 15.6 & 60.6 & 49.6 & 83.0 & 70.1 & 63.0 & 52.1 \\
$L=3$ (2$\times$) & 27.9 & 19.4 & 64.7 & 53.8 & 84.0 & 71.8 & 68.9 & 55.8 \\
$L=5$ (3$\times$) & 26.9 & 22.7 & 64.8 & \textbf{55.7} & 85.4 & 73.0 & 68.8 & 56.8 \\
$L=7$ (4$\times$) & \textbf{30.2} & \textbf{24.6} & \textbf{67.5} & 52.9 & 84.2 & \textbf{75.3} & \textbf{69.8} & \textbf{57.8} \\
$L=9$ (5$\times$) & 28.5 & 22.5 & 67.0 & 54.7 & 84.8 & 73.3 & 66.0 & 56.7 \\
\padbelow $L=11$ (6$\times$) & 29.2 & 23.5 & 65.8 & 55.3 & \textbf{85.8} & 71.3 & 66.8 & 56.8 \\
\end{tabular}
\end{table}

\paragraph{Llama-3.2-1B-Base and Qwen3.5-2B-Base.}
\cref{tab:app_llama} reports the per-benchmark scores of \cref{tab:llama}.

\begin{table}[H]
\caption{\textbf{Per-benchmark scores across model families.} We compare \method{} with SFT on Llama-3.2-1B-Base and Qwen3.5-2B-Base at 2$\times$ compute. Both code benchmarks report the plus tests. Subscripts give the change of \method{} relative to SFT in points. \emph{avg.} averages the seven math and code scores.}
\label{tab:app_llama}
\vspace{-7pt}
\newlength{\llamanum}\settowidth{\llamanum}{\footnotesize\textbf{00.0}}
\newlength{\llamasub}\settowidth{\llamasub}{\footnotesize\textsubscript{+00.0}}
\newcommand{\sft}[1]{\makebox[\llamanum][r]{#1}}
\newcommand{\ours}[2]{\makebox[\llamanum][r]{\textbf{#1}}\makebox[\llamasub][l]{\textsubscript{\better{#2}}}}
\tablestyle{7.5pt}{1.2}
\begin{tabular}{l|c>{\columncolor{baselinecolor}}cc>{\columncolor{baselinecolor}}c}
\padbelow\headspan{benchmark} & \multicolumn{2}{c}{Llama-3.2-1B-Base} & \multicolumn{2}{c}{Qwen3.5-2B-Base} \\
\cmidrule(lr){2-3}\cmidrule(lr){4-5}
\padabove\padbelow& SFT & \cellcolor{white}\textbf{\method{}} & SFT & \cellcolor{white}\textbf{\method{}} \\
\noalign{\vspace{1pt}}
\hline
\gaprow{\multicolumn{1}{c}{} & & & & }
\padabove AIME25 & \sft{0.4} & \ours{1.0}{+0.6} & \sft{24.6} & \ours{37.7}{+13.1} \\
AIME26 & \sft{0.2} & \ours{0.4}{+0.2} & \sft{24.0} & \ours{41.2}{+17.2} \\
AMC23 & \sft{13.3} & \ours{22.7}{+9.4} & \sft{64.7} & \ours{80.9}{+16.2} \\
OlympiadBench & \sft{9.5} & \ours{16.6}{+7.1} & \sft{52.0} & \ours{63.6}{+11.6} \\
\padbelow MATH500 & \sft{28.4} & \ours{43.8}{+15.4} & \sft{85.6} & \ours{89.6}{+4.0} \\
\gaprow{\multicolumn{1}{c}{} & & & & }
\hline
\gaprow{\multicolumn{1}{c}{} & & & & }
\padabove HumanEval+ & \sft{29.1} & \ours{32.5}{+3.4} & \sft{62.0} & \ours{71.0}{+9.0} \\
\padbelow MBPP+ & \sft{18.9} & \ours{20.5}{+1.6} & \sft{51.6} & \ours{64.8}{+13.2} \\
\gaprow{\multicolumn{1}{c}{} & & & & }
\hline
\gaprow{\multicolumn{1}{c}{} & & & & }
\padabove\padbelow avg. & \sft{14.3} & \ours{19.6}{+5.4} & \sft{52.1} & \ours{64.1}{+12.0} \\
\gaprow{\multicolumn{1}{c}{} & & & & }
\end{tabular}
\end{table}

\subsection{Validation Loss over Finetuning}
\label{app:loss}
\begin{figure}[H]
  \centering
  \includegraphics[width=\textwidth]{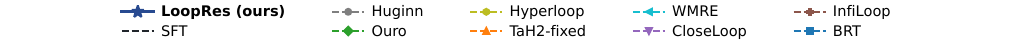}\\[2pt]
  \subcaptionbox{Loop of 2$\times$ compute.\label{fig:loss_ut2}}{%
    \includegraphics[width=0.49\textwidth]{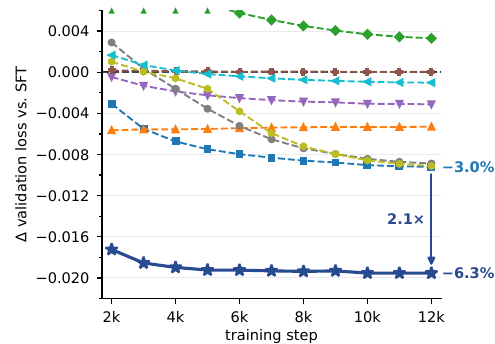}}\hfill
  \subcaptionbox{Loop of 4$\times$ compute.\label{fig:loss_ut4}}{%
    \includegraphics[width=0.49\textwidth]{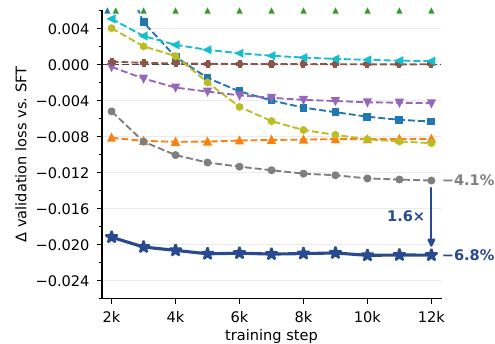}}
  \vspace{-3pt}
  \caption{\textbf{Validation loss relative to SFT.}
    We finetune Qwen3-1.7B-Base on code and math with each looped design and subtract the loss of the same model finetuned without looping (SFT), each point averaging 1,000 steps.
    Panels (a) and (b) give every design the compute of two and four full passes through the model, respectively.
    Triangles on the top edge mark values above the plotted range.}
  \label{fig:loss}
\end{figure}

\cref{fig:loss} reports the validation loss of each design relative to SFT over finetuning.

\subsection{Test-Time Scaling on More Backbones}
\label{app:tts}
\begin{figure}[H]
  \centering
  \subcaptionbox{Qwen3-30B-A3B-Base.\label{fig:scale30b}}{%
    \includegraphics[width=0.32\textwidth]{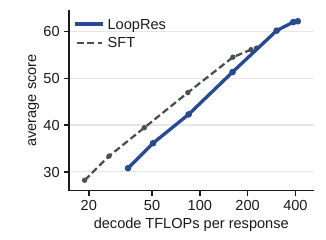}}\hfill
  \subcaptionbox{Llama-3.2-1B-Base.\label{fig:scalellama}}{%
    \includegraphics[width=0.32\textwidth]{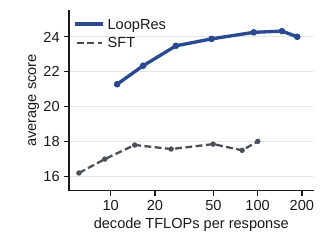}}\hfill
  \subcaptionbox{Qwen3.5-2B-Base.\label{fig:scaleqwen35}}{%
    \includegraphics[width=0.32\textwidth]{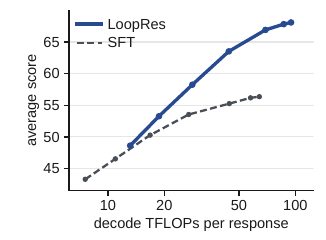}}
  \vspace{-3pt}
  \caption{\textbf{Test-time scaling on other backbones.}
    Average score against mean decode compute, as the thinking budget grows from 1K to 30K tokens.
    (a) averages the five benchmarks of \cref{tab:scale}, and (b) and (c) the seven of \cref{tab:app_llama}.
    \method{} runs at 2$\times$ compute, so it spends twice SFT's compute per token.}
  \label{fig:scale_backbones}
\end{figure}

\cref{fig:scale_backbones} reports the test-time scaling of \method{} and SFT on Qwen3-30B-A3B-Base, Llama-3.2-1B-Base, and Qwen3.5-2B-Base as the thinking budget grows from 1K to 30K tokens.

\subsection{Performance under Equal Token Budgets}
\label{app:tokens}
\begin{figure}[H]
  \centering
  \includegraphics[width=\textwidth]{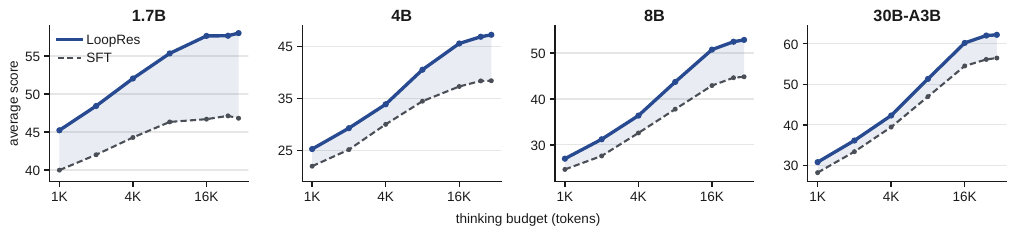}
  \vspace{-13pt}
  \caption{\textbf{Test-time scaling at equal token budgets.}
    Average score against the thinking budget, as it grows from 1K to 30K tokens.
    The 1.7B panel averages the seven benchmarks of \cref{tab:main} and the other three the five of \cref{tab:scale}.
    \method{} runs at 2$\times$ compute, so it spends twice SFT's compute per token, and the shaded band is its gain over SFT at the same budget.}
  \label{fig:scale_tokens}
\end{figure}

The figures above compare \method{} and SFT at equal decode compute, where \method{} spends twice SFT's compute per token.
\cref{fig:scale_tokens} instead reports both at equal token budgets on Qwen3-\{1.7B, 4B, 8B, 30B-A3B\}-Base.

\section{Additional Experiments}
\label{app:extra}

\subsection{Alternative Ways of Looping with AttnRes}
\label{app:huginn_all}
We run Huginn + AttnRes with all layers looped in place of the prelude, core, and coda, at $L=2$ so that it matches the 2$\times$ compute of the split version at $L=3$.
\cref{tab:app_huginn} reports the results.

\subsection{Applying AttnRes without Looping}
\label{app:attnres_noloop}
\method{} loops the core and reads its earlier loops with AttnRes, and we also apply AttnRes to the backbone without looping it, over the $B$ blocks of the single pass, with the rest of the recipe following the SFT run of \cref{tab:main}.
\cref{tab:app_attnres_noloop} reports the results.

\begin{table}[H]
  \centering
\begin{minipage}[t]{0.48\textwidth}
\caption{\textbf{Huginn variants.} Both variants run at 2$\times$ compute on Qwen3-1.7B-Base. \emph{core} loops the core of a prelude, core, and coda, as in \cref{tab:main,tab:ablation}, and \emph{all} loops the whole model. \emph{avg.} averages the seven math and code scores.}
\label{tab:app_huginn}
\vspace{-7pt}
\tablestyle{9pt}{1.2}
\begin{tabular}{l|cc|c}
\padabove\padbelow variant & looped layers & $L$ & avg. \\
\shline
\padabove Huginn & core & 3 & 49.0 \\
Huginn + AttnRes & all & 2 & 50.9 \\
\padbelow Huginn + AttnRes & core & 3 & \textbf{53.9} \\
\end{tabular}
\end{minipage}%
\hfill
\begin{minipage}[t]{0.48\textwidth}
\caption{\textbf{AttnRes without looping.} We compare SFT, SFT with AttnRes, and \method{} on Qwen3-1.7B-Base. The SFT and \method{} rows are those of \cref{tab:main}, and SFT + AttnRes applies block AttnRes to the $B$ blocks of the single pass. \emph{avg.} averages the seven math and code scores.}
\label{tab:app_attnres_noloop}
\vspace{-7pt}
\tablestyle{9pt}{1.2}
\begin{tabular}{l|c|c}
\padabove\padbelow method & compute & avg. \\
\shline
\padabove SFT & 1$\times$ & 43.2 \\
SFT + AttnRes & 1$\times$ & 44.3 \\
\padbelow \method{} & 2$\times$ & \textbf{55.8} \\
\end{tabular}
\end{minipage}%

\end{table}

\subsection{Comparison with SFT at Matched Training Compute}
\label{app:data2x}
The runs of \cref{tab:main} finetune for 12,000 steps, and we also train SFT for 24,000 steps, which matches the training compute of \method{} at 2$\times$ per token, and read its final checkpoint at step 24,000.
\cref{tab:app_data2x} reports the per-benchmark scores, and \cref{fig:app_data2x} the test-time scaling of both SFT models.

\begin{table}[H]
\caption{\textbf{SFT at matched training compute.} We compare SFT on 1$\times$ and 2$\times$ data with \method{} on Qwen3-1.7B-Base. Every row is scored under the decoding of \cref{app:eval} with no thinking budget. SFT at 1$\times$ data is read from the final checkpoint at step 12,000 and at 2$\times$ data from the final checkpoint at step 24,000, both at 1$\times$ compute, and the \method{} rows are those of \cref{tab:main}, trained on 1$\times$ data. HE+ denotes HumanEval+, and both code benchmarks report the plus tests. \emph{avg.} averages the seven math and code scores, and $\Delta$ gives the change relative to SFT at 1$\times$ data.}
\label{tab:app_data2x}
\vspace{-7pt}
\tablestyle{6.5pt}{1.2}
\begin{tabular}{l|ccccc|cc|cc}
\padbelow\headspan{setting} & \multicolumn{5}{c|}{math} & \multicolumn{2}{c|}{code} & \headspan{avg.} & \headspan{$\Delta$} \\
\cmidrule(lr){2-6}\cmidrule(lr){7-8}
\padabove\padbelow& AIME25 & AIME26 & AMC23 & Olympiad & MATH500 & HE+ & MBPP+ & & \\
\shline
\padabove SFT, 1$\times$ data & 14.6 & 11.7 & 48.6 & 40.7 & 73.4 & 57.3 & 56.2 & 43.2 & -- \\
\padbelow SFT, 2$\times$ data & 20.8 & 15.2 & 56.9 & 45.9 & 80.2 & 66.6 & 60.7 & 49.5 & \better{+6.3} \\
\shline
\padabove \method{}, 2$\times$ compute & 27.9 & 19.4 & 64.7 & 53.8 & 84.0 & 71.8 & 68.9 & 55.8 & \better{+12.6} \\
\padbelow \method{}, 4$\times$ compute & 30.2 & 24.6 & 67.5 & 52.9 & 84.2 & 75.3 & 69.8 & 57.8 & \better{+14.6} \\
\end{tabular}
\end{table}

\subsection{Looping a 4B Model against a Non-Looped 8B Model}
\label{app:4b8b}
\method{} doubles the compute of a backbone without adding parameters, and we compare it against a backbone of twice the parameters at the same decode compute.
\cref{fig:scale4b8b} reports the test-time scaling of \method{} on Qwen3-4B-Base and SFT on Qwen3-8B-Base as the thinking budget grows from 1K to 30K tokens.

\begin{figure}[H]
  \centering
  \input{figures_tex/app_data2x}\hfill
  \input{figures_tex/scaling_4b8b}
\end{figure}

\end{document}